\documentclass{article} % For LaTeX2e
\usepackage{hyperref}
\usepackage{url}

\usepackage[backend=bibtex,style=numeric,natbib=true,sorting=none,doi=false,url=false,maxbibnames=99,firstinits=true]{biblatex}
\usepackage{times}
\usepackage{graphicx} % more modern                                                                                  
\usepackage{rotating}
\usepackage{caption}
\usepackage{subcaption}
\usepackage{multirow}
\usepackage{ctable} % for \specialrule command

\usepackage[font=small,labelfont=bf]{caption}
\usepackage{algorithm}
\usepackage{algorithmic}
\usepackage{dsfont}
\usepackage{amsmath}
\usepackage{amsfonts}
\usepackage{amssymb}
\usepackage{authblk}
\usepackage{enumitem}
\usepackage{booktabs}

\definecolor{gray}{rgb}{.7, .7, .7}

\newcommand{\sminus}{\scalebox{0.75}[1.0]{-}}

\title{Parallel Multi-Dimensional LSTM, With Application to Fast Biomedical Volumetric Image Segmentation}

\author[*123]{Marijn F. Stollenga}
\author[*1245]{Wonmin Byeon}
\author[4]{Marcus Liwicki}
\author[123]{Juergen Schmidhuber}
\affil[*]{\small{Shared first authors, both Authors contributed equally to this work. Corresponding authors: \texttt{marijn@idsia.ch}, \texttt{wonmin.byeon@dfki.de}}}
\affil[1]{Istituto Dalle Molle di Studi sull'Intelligenza Artificiale (The Swiss AI Lab IDSIA)}
\affil[2]{Scuola universitaria professionale della Svizzera italiana (SUPSI), Switzerland}
\affil[3]{Universit\'{a} della Svizzera italiana (USI), Switzerland}
\affil[4]{University of Kaiserslautern, Germany}
\affil[5]{German Research Center for Artificial Intelligence (DFKI), Germany}

\newcommand{\abr}{PyraMiD-LSTM}

\begin{document}

\maketitle

\begin{abstract}
Convolutional Neural Networks (CNNs) can be shifted across 2D images or 3D videos to segment them. They have a fixed input size and typically perceive only small local contexts of the pixels to be classified as foreground or background. In contrast, Multi-Dimensional Recurrent NNs (MD-RNNs) can perceive the entire spatio-temporal context of each pixel in a few sweeps through all pixels, especially when the RNN is a Long Short-Term Memory (LSTM). Despite these theoretical advantages, however, unlike CNNs, previous MD-LSTM variants were hard to parallelize on GPUs. Here we re-arrange the traditional cuboid order of computations in MD-LSTM in pyramidal fashion. The resulting PyraMiD-LSTM is easy to parallelize, especially for 3D data such as stacks of brain slice images. PyraMiD-LSTM achieved best known pixel-wise brain image segmentation results on MRBrainS13 (and competitive results on EM-ISBI12).
\end{abstract}

%juergen I liked the flow of my old abstract better - improved it above -  the stuff below somehow tries to say more but fails:
%Multi-Dimensional Recurrent Neural Networks (MD-RNNs) can perceive and effectively utilize the spatial context of image parts in the image segmentation task, especially when the recurrent network is the Long Short-Term Memory (MD-LSTM), by treating 2D image or 3D video as a multi-dimensional signal. In contrast, recently popular Convolutional Neural Network classifiers have a fixed input size and typically perceive only small local contexts of images or videos across which they have to be shifted for purposes of segmentation. Despite these theoretical advantages, however, up-to-date MD-LSTM variants were limited in parallelization on GPUs to i.e. parallelization within the simplex planes, where the context from the orthogonal directions is used by the recurrent layer neurons.  Here, this traditional order of computation is re-arranged such that the simplex planes turn into orthogonal planes that remain of constant size across the whole input. The context information flows in pyramids rather than cuboids, hence \abr{}. The method achieves efficient parallelization, especially for 3D data such as videos or stacks of images of brain slices. Experimental results have shown competitive results in pixel-wise segmentation on EM-ISBI12 and superior results on MRBrainS13, both brain segmentation datasets.

%juergen as always, missing references are in my public bib file http://people.idsia.ch/~juergen/bib.bib

\section{Introduction}
Long Short-Term Memory (LSTM) networks~\cite{lstm97and95,Gers:99c} are recurrent neural networks (RNNs) initially designed for sequence processing. They
achieved state-of-the-art results on challenging tasks such as
handwriting recognition~\cite{Graves:09tpami},
large vocabulary speech recognition~\cite{sak2014,sak2014large} and
machine translation~\citep{sutskever2014}.
Their architecture contains gates to store and read out information 
from linear units called error carousels that retain information over long time intervals, 
which is hard for traditional RNNs.

Multi-Dimensional LSTM networks (MD-LSTM~\cite{DBLP:conf/icann/GravesFS07}) connect hidden LSTM units in grid-like fashion\footnote{For example, in two dimensions this yields 4 directions; up, down, left and right.}.
Two dimensional MD-LSTM is applicable to image segmentation~\cite{DBLP:conf/icann/GravesFS07,graves:2009nips,byeon15}
where each pixel is assigned to a class such as background or foreground. 
Each LSTM unit sees a pixel and receives input from neighboring LSTM units, 
thus recursively gathering information about all other pixels in the image.

There are many biomedical 3D volumetric data sources, 
such as computed tomography (CT), magnetic resonance (MR), and electron microscopy (EM). 
Most previous approaches process each 2D slice separately, using image segmentation algorithms such as snakes~\cite{Kass88}, random forests~\cite{Wang15} and Convolutional Neural Networks~\cite{ciresan2012nips}. 3D-LSTM, however, can  process the full context of each pixel in such a volume through 8 sweeps over all pixels by 8 different LSTMs, each sweep in the general direction of one of the 8 directed volume diagonals. 

Due to the sequential nature of RNNs, however, MD-LSTM parallelization was difficult, especially for volumetric data.
The novel 
Pyramidal Multi-Dimensional LSTM (\abr{}) networks introduced in this paper use a rather different topology and update strategy. 
They are easier to parallelize, need fewer computations overall, and scale well on GPU architectures.

\abr{} is applied to two challenging tasks involving segmentation of biological volumetric images. Competitive results are achieved on EM-ISBI12~\cite{cardona10}; best known results are achieved on MRBrainS13~\cite{MICCAI15}.

%biomedical image segmentation (3D volume)
%limitation of CNN
%MDLSTM for GPU?

\section{Method}
\label{sec:method}
We will first describe standard one-dimensional LSTM~\cite{Gers:99c} and MD-LSTM. Then we introduce topology changes to construct the \abr{}, which is formally described and discussed.

The original LSTM unit consists of an input gate ($i$), forget gate\footnote{Although the forget gate output is inverted and actually `remembers' when it is on, and forgets when it is off, the traditional nomenclature is kept.} ($f$), output gate ($o$), and memory cell ($c$) which control what should be remembered or forgotten over potentially long periods of time. All gates and activations are real-valued vectors: $x, i, f, \tilde{c}, c, o, h \in \mathds{R}^T$, where $T$ is the length of the input. The gates and activations at discrete time $t$ ($t$=1,2,...) are computed as follows:
\begin{eqnarray}
i_t = \sigma(x_t \cdot \theta_{xi} + h_{t \sminus 1} \cdot \theta_{hi} + \theta_{ibias}),\\
f_t = \sigma(x_t \cdot \theta_{xf} + h_{t \sminus 1} \cdot \theta_{hf} + \theta_{fbias}),\\
\tilde{c}_t = \tanh(x_t \cdot \theta_{x\tilde{c}} + h_{t\sminus 1} \cdot \theta_{h\tilde{c}} + \theta_{\tilde{c}bias}),\\
c_t = \tilde{c}_t \odot i_t + c_{t \sminus 1} \odot f_t,\\
o_t = \sigma(x_t \cdot \theta_{xo} + h_{t \sminus 1} \cdot \theta_{ho} + \theta_{obias}),\\
h_t = o_t \odot \tanh(c_t)
\end{eqnarray}
where ($\cdot$) is a matrix multiplication, ($\odot$) an element-wise multiplication, and $\theta$ denotes the weights. 
%$i$, $f$ and $o$ are the input-, forget-, and output gates respectively\footnote{Although the forget gate actually `remembers' when it is on, and forgets when it is off, we keep this naming since it is traditional.}.
$\tilde{c}$ is the input to the 'cell' $c$, which is gated by the input gate, and $h$ is the output.
The non-linear functions $\sigma$ and $\tanh$ are applied element-wise, where $\sigma(x) = \frac{1}{1 + e^{-x}}$.
Equations (1, 2) determine gate activations, Equation (3) cell inputs,
Equation (4) the new cell states (here `memories' are stored or forgotten),
Equation (5) output gate activations which appear in Equation (6), the final output.

\subsection{Pyramidal Connection Topology}

\begin{figure}[h]
\centering
\includegraphics[width=12.0cm]{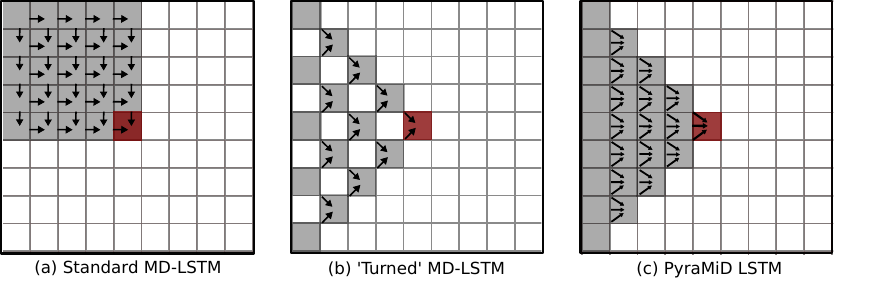}
\caption{The standard MD-LSTM topology (a) evaluates the context of each pixel recursively from neighboring pixel contexts along the axes, that is, pixels on a simplex can be processed in parallel. Turning this order by $45^\circ$ (b) causes the simplex to become a plane (a column vector in the 2D case here). The resulting gaps are filled by adding extra connections, to process more than 2 elements of the context (c).}
%juergen careful here - we don't process pixels, but recursively process neighboring pixel contexts
\label{fig:context}
\end{figure}

In MD-LSTMs, connections are aligned with the grid axes. 
In 2D, these directions are up, down, left and right. 
A 2D-LSTM adds the pixel-wise outputs of 4 LSTMs, 
one scanning the image pixel by pixel from north-west to south-east,
one from north-east to south-west,
one from south-west to north-east,
and one from south-east to north-west.

If the connections are rotated by $45^\circ$, all inputs to all units come from either left, right, up, or down (left in case of Figure~\ref{fig:context}--b).
This greatly facilitates parallelization, 
since all the elements of a whole grid row can be computed independently, 
which does not work for MD-LSTM simplexes, whose sizes vary.
However, this introduces context gaps as in Figure~\ref{fig:context}--b.
By adding an extra input, these gaps are filled as  in Figure~\ref{fig:context}--c.

A similar connection strategy has been previously used to speed up non-euclidian distance computations on surfaces~\cite{weber2008parallel}. 
There are however important differences:
\begin{itemize}[noitemsep]
\item We can exploit efficient GPU-based CUDA convolution operations, but in a way unlike what is done in CNNs, as will be explained below.
\item As a result of these operations, input filters that are bigger than the necessary $3\times3$ filters arise naturally, creating overlapping contexts. Such redundancy turns out to be beneficial and is used in our experiments.
\item We apply several layers of complex processing with multi-channeled outputs and several state-variables for each pixel, instead of having a single value per pixel as in distance computations.
\item Our application is focused on volumetric data.
\end{itemize}

\begin{figure}[h]
\centering
\includegraphics[width=6.0cm]{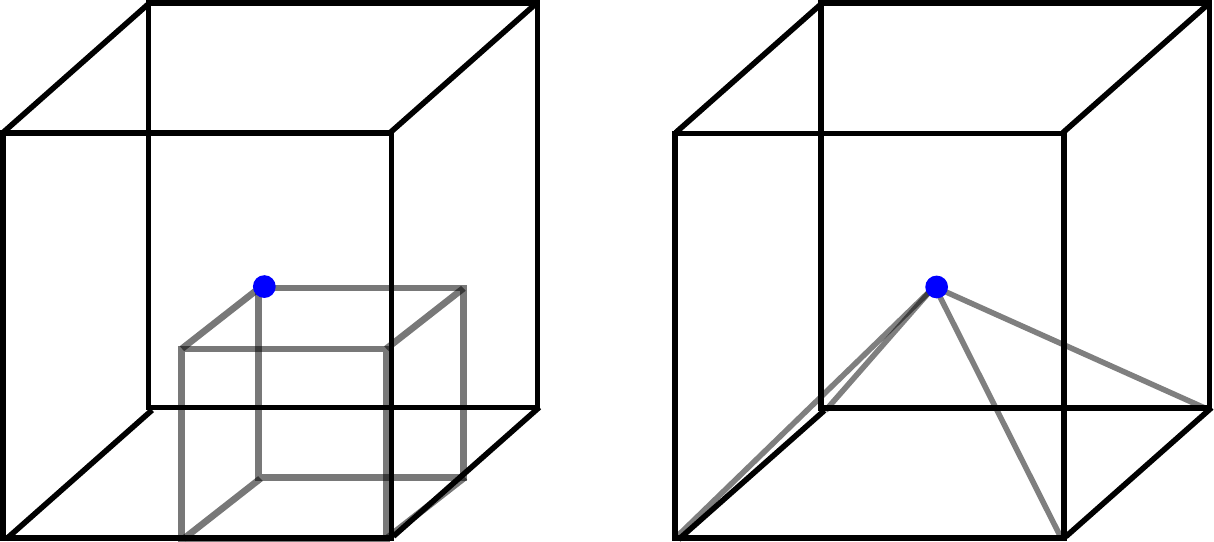}
\caption{On the \textbf{left} we see the  context scanned so far by one of the 8 LSTMs of a 3D-LSTM: a cube. In general, given $d$ dimensions, $2^d$ LSTMs are needed.
On the \textbf{right} we see the context scanned so far by one of the 6 LSTMs of a 3D-\abr{}: a pyramid. In general, $2 \times d$ LSTMs are needed.}
\label{fig:cubes}
\end{figure}

One of the striking differences between \abr{} and MD-LSTM is the shape of the scanned contexts.
Each LSTM of an MD-LSTM scans rectangle-like contexts in 2D or cuboids in 3D.
Each LSTM of a \abr{} scans triangles in 2D and pyramids in 3D (see Figure~\ref{fig:cubes}).
An MD-LSTM needs 8 LSTMs to scan a volume, while a \abr{} needs only 6, since it takes 8 cubes or 6 pyramids to fill a volume. 
Given dimension $d$, the number of LSTMs grows as $2^d$ for an MD-LSTM (\emph{exponentially}) and $2\times d$ for a \abr{} (\emph{linearly}).

\subsection{\abr{}}

% JAN ended his editing here

\begin{figure}[h]
\centering
\includegraphics[width=\textwidth]{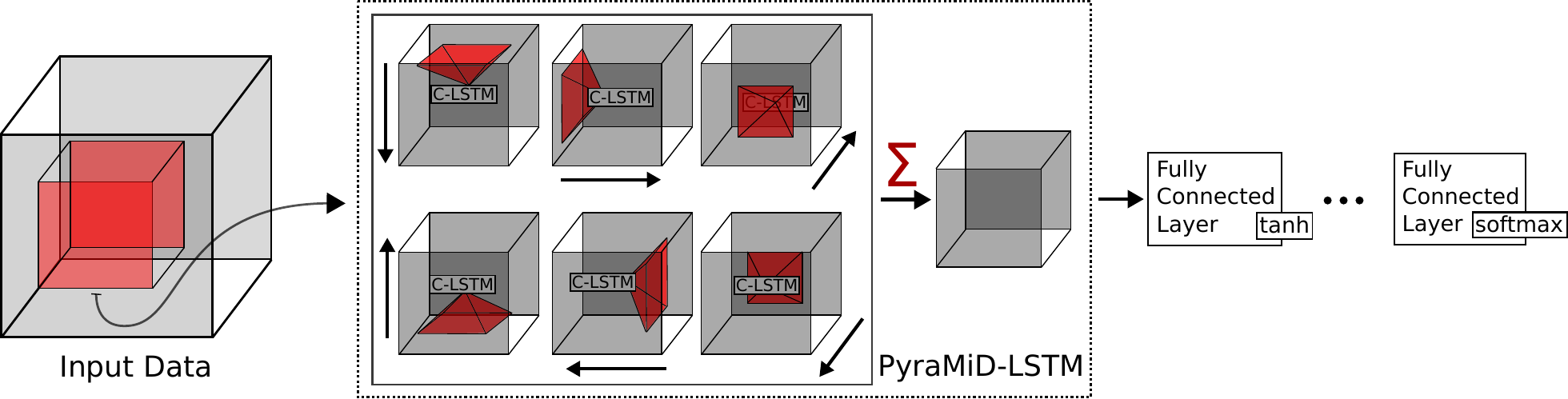}
\caption{\abr{} network architecture. Randomly rotated and flipped inputs are sampled from random locations, then fed to six C-LSTMs over three axes. The outputs from all C-LSTMs are combined into one \abr{} layer and sent to the fully-connected layer. $tanh$ is used as a squashing function in the hidden layer. Several \abr{} layers can be applied. The last layer is fully-connected and uses a softmax function to compute probabilities for each class for each pixel.}
%juergen here you mean by layers entire LSTM RNNs ... better call them thus - they are not normal layers ...
\label{fig:architecture}
\end{figure}

Here we explain the \abr{} network architecture for 3D volumes (see Figure~\ref{fig:architecture}). It consists of six LSTMs with RNN-tailored convolutions (C-LSTM), one for each direction, to create the full context of each pixel.
Note that each of these C-LSTMs is a entire LSTM RNN, processing the entire volume in one direction.
The directions $\mathcal{D}$ are defined over the three axes $(x, y, z)$:  $\mathcal{D} = \{(\cdot, \cdot, 1), (\cdot, \cdot, -1), (\cdot, 1, \cdot), (\cdot, -1, \cdot), \linebreak[0](1, \cdot, \cdot), (-1, \cdot, \cdot)\}$.

Each C-LSTM performs computations in a plane moving in the defined direction.
The symbol ($\cdot$) in a dimension signifies that the plane is parallel to this axis, 
and a $1$ or $-1$ implies that the computation is moving along the positive or negative direction of that axis, respectively.
The input is $x \in \mathds{R}^{W \times H \times D \times C}$, where $W$ is the width, $H$ the height, $D$ the depth, and $C$ the number of channels of the input, or hidden units in the case of second- and higher layers. Similarly, we define the volumes $f^d, i^d, o^d, \tilde{c}^d, c^d, h^d, h \in \mathds{R}^{W \times H \times D \times O}$, where $d \in \mathcal{D}$ is a direction and $O$ is the number of hidden units (per pixel).
Since each direction needs a separate volume, we denote volumes with $(\cdot)^d$. 

To keep the equations readable, we omit the time index below: $h_t$ becomes $h$ and $h_{t-1}$ becomes $h_{\sminus 1}$.
The time index $t$ is bound to the axis along which computations are performed.
For instance, for direction $d = (\cdot, \cdot, 1)$, $v^d$ refers to the plane orthogonal to axis z; i.e. $v_{x, y, z, c}$ for $x=1..X, y=1..Y, c=1..C$, and $z = t$.
For a negative direction $d = (\cdot, \cdot, -1)$, the plane is the same but moves in the opposite direction: $z = Z - t$.
Furthermore, $v^{d}_{\sminus 1}$ refers to the previous plane, in this case  $v_{x, y, z, c}$ for $x=1..X, y=1..Y, c=1..C, z=t-1$.
A special case is the first plane in each direction, which does not have a previous plane, hence we omit the corresponding computation.
The following functions are defined for all planes and all directions:

\vspace{0.5em}
\textbf{C-LSTM:}
\begin{eqnarray}
i^d = \sigma(x \ast \theta_{xi}^d + h^{d}_{\sminus 1} \ast \theta^d_{hi} + \theta^d_{ibias}),\\
f^d = \sigma(x \ast \theta_{xf}^d + h^{d}_{\sminus 1} \ast \theta^d_{hf} + \theta^d_{fbias}),\\
\tilde{c}^d = \tanh(x \ast \theta_{x\tilde{c}}^d + h^{d}_{\sminus 1} \ast \theta^d_{h\tilde{c}} + \theta^d_{\tilde{c}bias}),\\
c^d = \tilde{c}^d \odot i^d + c^{d}_{\sminus 1} \odot f^d,\\
o^d = \sigma(x \ast \theta_{xo}^d + h^{d}_{\sminus 1} \ast \theta^d_{ho} + \theta^d_{obias}),\\
h^d = o^d \odot \tanh(c^d),
\end{eqnarray}

where ($\ast$) is a RNN-tailored convolution\footnote{In 3D volumes, convolutions are performed in 2D; in general an n-D volume requires n-1-D convolutions. All convolutions have stride 1, and their filter sizes should at least be $3\times 3$ in each dimension to create the full context.}, and $h$ is the output of the layer. Note that in C-LSTMs the convolution operations propagate information `sideways' \emph{over the axis of the data}. This is very different from CNNs where the convolution operations propagate information upwards to the next layer. All biases are the same for all LSTM units (i.e., no positional biases are used). The outputs $h^d$ for all directions are added, combining all C-LSTMs into one \abr{}:
%h^d = \tanh(x \ast \theta_{xh}^d + c_d \ast \theta^d_{ch} + \theta^d_{hbias}),\\
\begin{eqnarray}
h = \sum_{d\in \mathcal{D}} h^d.
\end{eqnarray}

%juergen above convolutions appear - you want to keep this?

\textbf{Fully-Connected Layer:}
Each \abr{} layer is connected to a fully-connected layer, and the output is squashed by the hyperbolic tangent ($tanh$) function. At the end, a softmax function is applied after the last fully-connected layer.
%Figure~\ref{fig:architecture} illustrates our network architecture. It consists of CV-LSTM, fully connected layers and the output layer with a softmax function.

%\section{Related Work}

\section{Experiments}
\label{sec:exp}
We evaluate our approach on two 3D biomedical image segmentation datasets: electron microscopy (EM) and MR Brain images.

\paragraph{EM dataset} 
The EM dataset~\cite{cardona10} is provided by the ISBI 2012 workshop on Segmentation of Neuronal Structures in EM Stacks \cite{isbi12}. Two stacks consist of 30 slices of $512\times512$ pixels obtained from a $2 \times 2 \times\ 1.5\;\mu m^3$ microcube with a resolution of $4\times4\times50$ $nm^3$/pixel and binary labels. One stack is used for training, the other for testing. Target data consists of binary labels (membrane and non-membrane). 

\paragraph{MR Brain dataset}
The MR Brain images are provided by the ISBI 2015 workshop on Neonatal and Adult MR Brain Image Segmentation (ISBI NEATBrainS15)~\cite{MICCAI15}. The dataset consists of twenty fully annotated high-field (3T) multi-sequences: 3D T1-weighted scan (T1), T1-weighted inversion recovery scan (IR), and fluid-attenuated inversion recovery scan (FLAIR). The dataset is divided into a training set with five volumes and a test set with fifteen volumes. All scans are bias-corrected and aligned. Each volume includes 48 slices with $240\times240$ pixels ($3mm$ slice thickness). The slices are manually segmented through nine labels: cortical gray matter, basal ganglia, white matter, white matter lesions, cerebrospinal fluid in the extracerebral space, ventricles, cerebellum, brainstem, and background. Following the ISBI NEATBrainS15 workshop procedure, all labels are grouped into four classes and background: 1) cortical gray matter and basal ganglia (GM), 2) white matter and white matter lesions (WM), 3) cerebrospinal fluid and ventricles (CSF), and 4) cerebellum and brainstem. Class 4) is ignored for the final evaluation as required.

\paragraph{Sub-volumes and Augmentation} 
The full dataset requires more than the 12 GB of memory provided by our GPU, hence we train and test on sub-volumes.
We randomly pick a position in the full data and extract a smaller cube (see the details in \textit{Bootstrapping}).
This cube is possibly rotated at a random angle over some axis and can be flipped over any axis.
For EM images, we rotate over the z-axis and flipped sub-volumes with 50\% chance along x, y, and z axes. 
For MR brain images, rotation is disabled; only flipping along the x direction is considered, 
since brains are (mostly) symmetric in this direction.

During test-time, rotations and flipping are disabled and the results of all sub-volumes are stitched together using a Gaussian kernel, providing the final result.

\paragraph{Pre-processing}
We normalize each input slice towards a mean of zero and variance of one, since the imaging methods sometimes yield large variability in contrast and brightness. We do not apply the complex pre-processing common in biomedical image segmentation~\cite{Wang15}. 

\begin{figure}[t!]
        \centering
        \begin{subfigure}[b]{0.32\textwidth}
                \includegraphics[width=\textwidth]{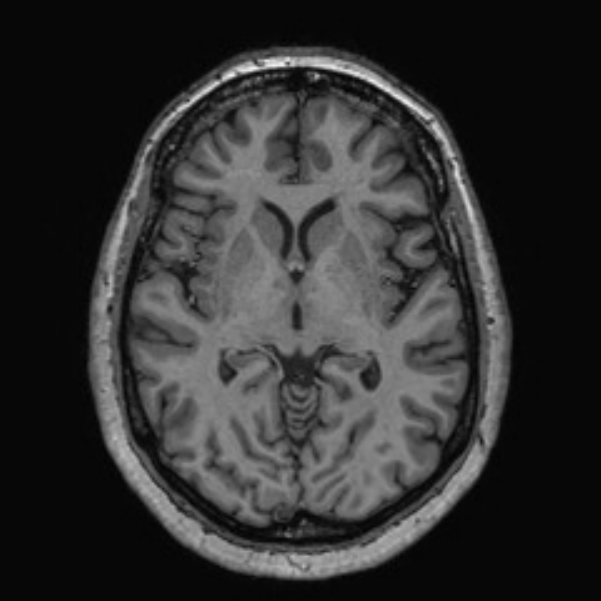}
                \caption{T1}
                \label{fig:T1}
        \end{subfigure}
        \begin{subfigure}[b]{0.32\textwidth}
                \includegraphics[width=\textwidth]{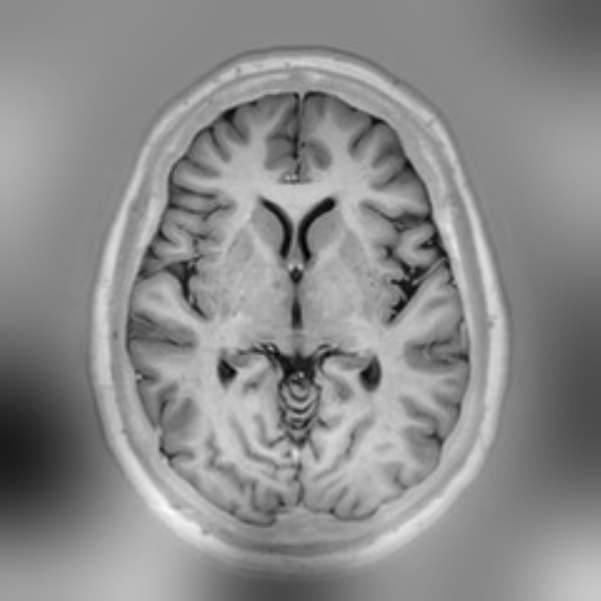}
                \caption{IR}
                \label{fig:IR}
        \end{subfigure}
        \begin{subfigure}[b]{0.32\textwidth}
                \includegraphics[width=\textwidth]{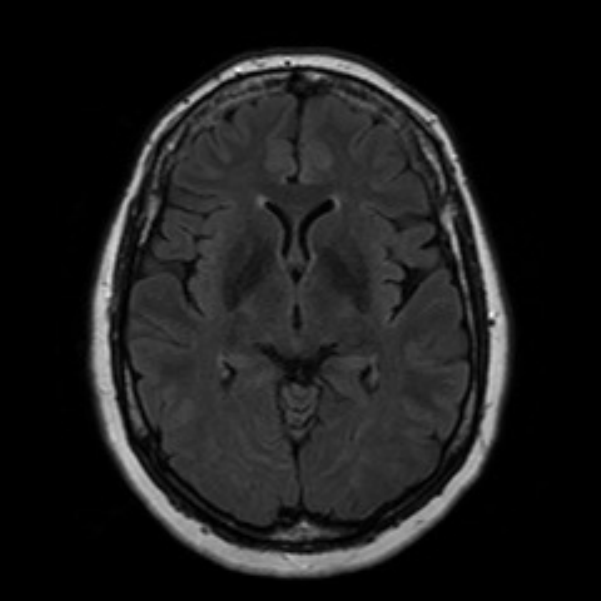}
                \caption{FLAIR}
                \label{fig:FLAIR}
        \end{subfigure}
        \\ \vspace{0.4em}
        \begin{subfigure}[b]{0.32\textwidth}
                \includegraphics[width=\textwidth]{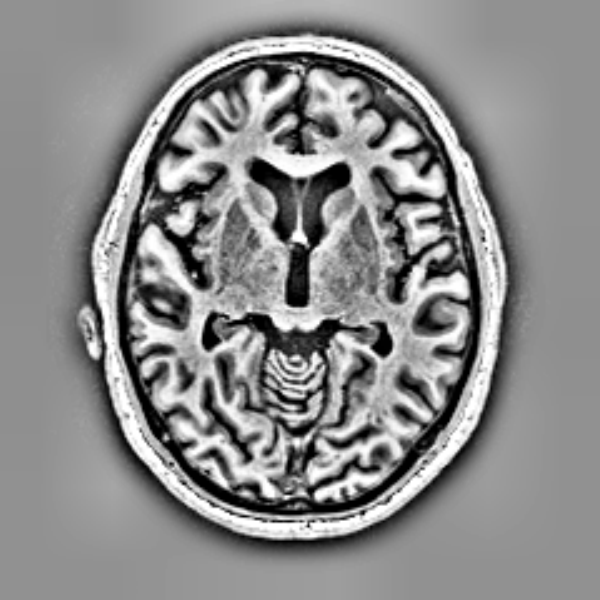}
                \caption{T1 (pre-processed)}
                \label{fig:T1}
        \end{subfigure}
        \begin{subfigure}[b]{0.32\textwidth}
                \includegraphics[width=\textwidth]{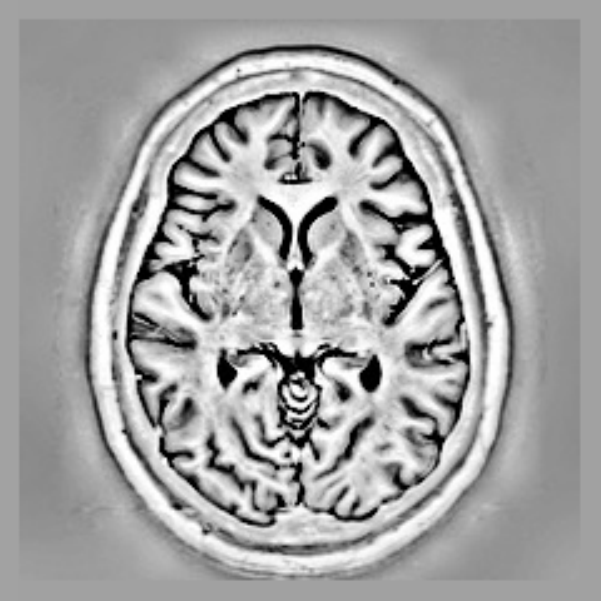}
                \caption{IR (pre-processed)}
                \label{fig:IR-PP}
        \end{subfigure}
        \begin{subfigure}[b]{0.32\textwidth}
                \includegraphics[width=\textwidth]{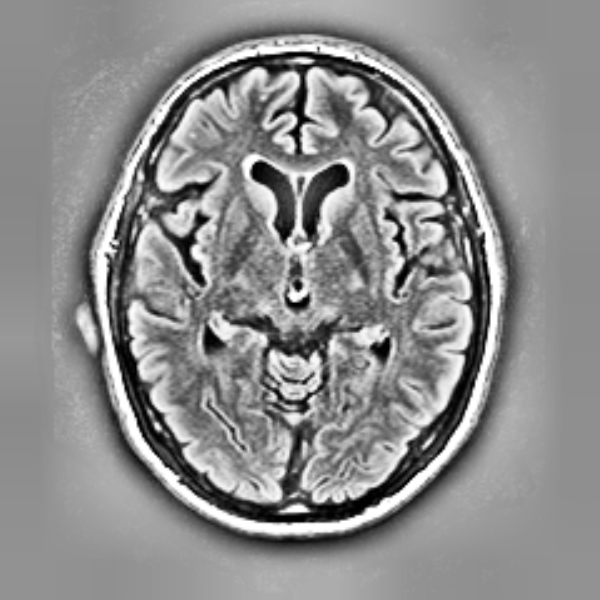}
                \caption{FLAIR (pre-processed)}
                \label{fig:FLAIR}
        \end{subfigure}
        \\ \vspace{2em}
        \begin{subfigure}[b]{0.48\textwidth}
                \includegraphics[width=\textwidth]{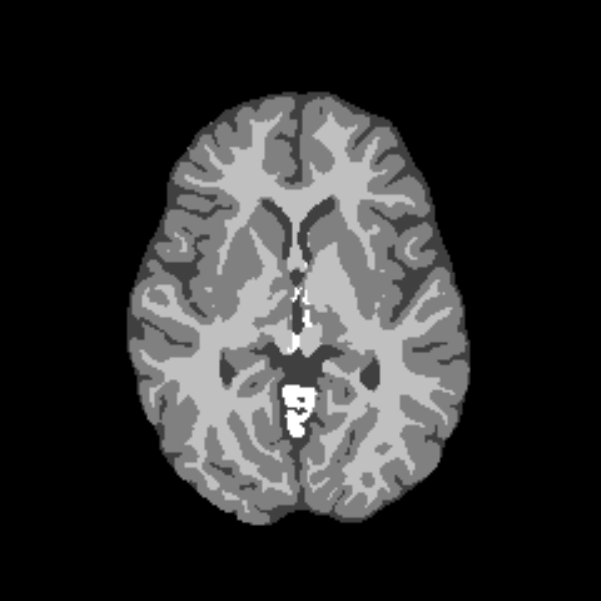}
                \caption{segmentation result from \abr{}}
                \label{fig:cvlstm}
        \end{subfigure}
        \caption{\textbf{Slice 19 of the test image 1.} (a)-(c) are examples of three scan methods used in the MR brain dataset, and (d)-(f) show the corresponding images after our pre-processing procedure (see pre-processing in Section \label{sec:exp}). Input (b) is omitted due to strong artifacts in the data --- the other datatypes are all used as input to the \abr{}. The segmentation result is shown in (g).}\label{fig:result-brain}
\end{figure}

We apply simple pre-processing on the three datatypes of the MR Brain dataset, 
since they contain large brightness changes under the same label (even within one slice; see Figure~\ref{fig:result-brain}).
From all slices we subtract the Gaussian smoothed images (filter size: $31\times31$, $\sigma=5.0$), then a Contrast-Limited Adaptive Histogram Equalization (CLAHE)~\cite{Pizer87} is applied to enhance the local contrast (tile size: $16\times16$, contrast limit: 2.0). An example of the images after pre-processing is shown in Figure~\ref{fig:result-brain}.
The original and pre-processed images are all used, except the original IR images (Figure~\ref{fig:IR}), which have high variability.

\paragraph{Training} We apply RMS-prop~\cite{tieleman2012lecture,dauphin2015rmsprop} with momentum.
We define $a \xleftarrow{\rho} b$ to be $a_n = \rho a_n + (1 - \rho) b_n$, where $a, b \in \mathds{R}^N$.
The following equations hold for every epoch:

\begin{eqnarray}
E = (y^* - y)^2,\\
MSE \xleftarrow{\rho_{MSE}} \nabla^2_\theta E,\\
G = \frac{\nabla_\theta E}{\sqrt{MSE}+\epsilon},\\
M \xleftarrow{\rho_{M}} G,\\
\theta = \theta - \lambda_{lr} M,
\end{eqnarray}
where $y^*$ is the target, $y$ is the output from the networks, $E$ is the squared loss, $MSE$ a running average of the variance of the gradient, $\nabla^2$ is the element-wise squared gradient, $G$ the normalized gradient, $M$ the smoothed gradient, and $\theta$ the weights.
This results in normalized gradients of similar size for all weights, such that even weights with small gradients get updated.
This also helps to deal with vanishing gradients \cite{Hochreiter:91}.

We use a decaying learning rate: $\lambda_{lr} = 10^{-6} + 10^{-2} \left(\sqrt[100]{\frac{1}{2}}\right)^{epoch}$, which starts at $\lambda_{lr} \approx 10^{-2}$ and halves every 100 epochs asymptotically towards $\lambda_{lr} = 10^{-6}$. Other hyper-parameters used are $\epsilon=10^{-5}$, $\rho_{MSE}=0.9$, and $\rho_{M}=0.9$.

\paragraph{Bootstrapping}
To speed up training, we run three learning procedures with increasing sub-volume sizes:
first, 3000 epochs with size $64 \times 64 \times 8$, then 2000 epochs with size $128 \times 128 \times 15$. Finally, for the EM-dataset, we train 1000 epochs with size $256 \times 256 \times 20$, and for the MR Brain dataset 1000 epochs with size $240 \times 240 \times 25$.
After each epoch, the learning rate $\lambda_{lr}$ is reset.

\paragraph{Experimental Setup}
All experiments are performed on a desktop computer with an NVIDIA GTX TITAN X 12GB GPU. 
For GPU implementation, the NVIDIA CUDA Deep Neural Network library (cuDNN)~\cite{cudnn14} is used. On the MR brain dataset, training took around three days, and testing per volume took around 2 minutes.

We use exactly the same hyper-parameters and architecture for both datasets.
Our networks contain three \abr{} layers. The first \abr{} layer has 16 hidden units followed by a fully-connected layer with 25 hidden units. In the next \abr{} layer, 32 hidden units are connected to a fully-connected layer with 45 hidden units. In the last \abr{} layer, 64 hidden units are connected to the fully-connected output layer whose size equals the number of classes. 

The convolutional filter size for all \abr{} layers is set to $7\times 7$.
The total number of weights is 10,751,549, and all weights are initialized according to a uniform distribution: $\mathcal{U}(-0.1, 0.1)$.

%\todo{system spec}
%\todo{parameters; filter sizes, hidden size}
%LSTM architecutre
%Time?
\subsection{Neuronal Membrane Segmentation}
%dataset
%The dataset was provided by the ISBI 2012 workshop on Segmentation of Neuronal Structures in EM Stacks \cite{isbi12}.
%three error metrics
\paragraph{Evaluation metrics} Three error metrics evaluate the following factors:

\begin{itemize}
\item Rand error~\cite{Rand1971}: 1 - F-score of rand index, which measures similarity between two segmentations on the foreground.
\item Warping error~\cite{Bollmann2010}: topological disagreements (object splits and mergers)
\item Pixel error: 1 - F-score of pixel similarity
\end{itemize}

%results
\paragraph{Results} 
\begin{table}[t]
\centering
\caption{Performance comparison on EM images. Some of the competing methods reported in the ISBI 2012 website are not yet published. Comparison details can be found under \url{http://brainiac2.mit.edu/isbi_challenge/leaders-board}. }
\begin{tabular}{l|ccc}
\toprule
Group    & Rand Err. & Warping Err.($\times10^{-3}$) & Pixel Err.\\
%    						 & ($10^{-1}$) & Warping Err. & Pixel Err.\\
%\hline \hline \\ [-1.7ex]
\midrule
Human     						& 0.002   	& 0.0053   	& 0.001   \\ 
Simple Thresholding     		& 0.450    	& 17.14  	& 0.225   \\
%\specialrule{.1em}{.05em}{.05em}
\midrule
IDSIA~\cite{ciresan2012nips}	& 0.050    	& 0.420   	& 0.061   \\
DIVE     					& 0.048   	& \textbf{0.374}   	& \textbf{0.058}   \\
%\hline
\textbf{\abr{}}     		& \textbf{0.047}    	& 0.462  	& 0.062   \\
%\specialrule{.1em}{.05em}{.05em}
\midrule
%SCI     & 0.028054308    & 0.000515747   & 0.063349324   \\
IDSIA-SCI     					& 0.0189   	& 0.617   	& 0.103   \\
DIVE-SCI    					& 0.0178   	& 0.307   	& 0.058   \\
\bottomrule
\end{tabular}
\label{tab:result-em}
\end{table}
\begin{figure}[h]
    		\centering
        \begin{subfigure}[b]{0.3\textwidth}
                \includegraphics[width=\textwidth]{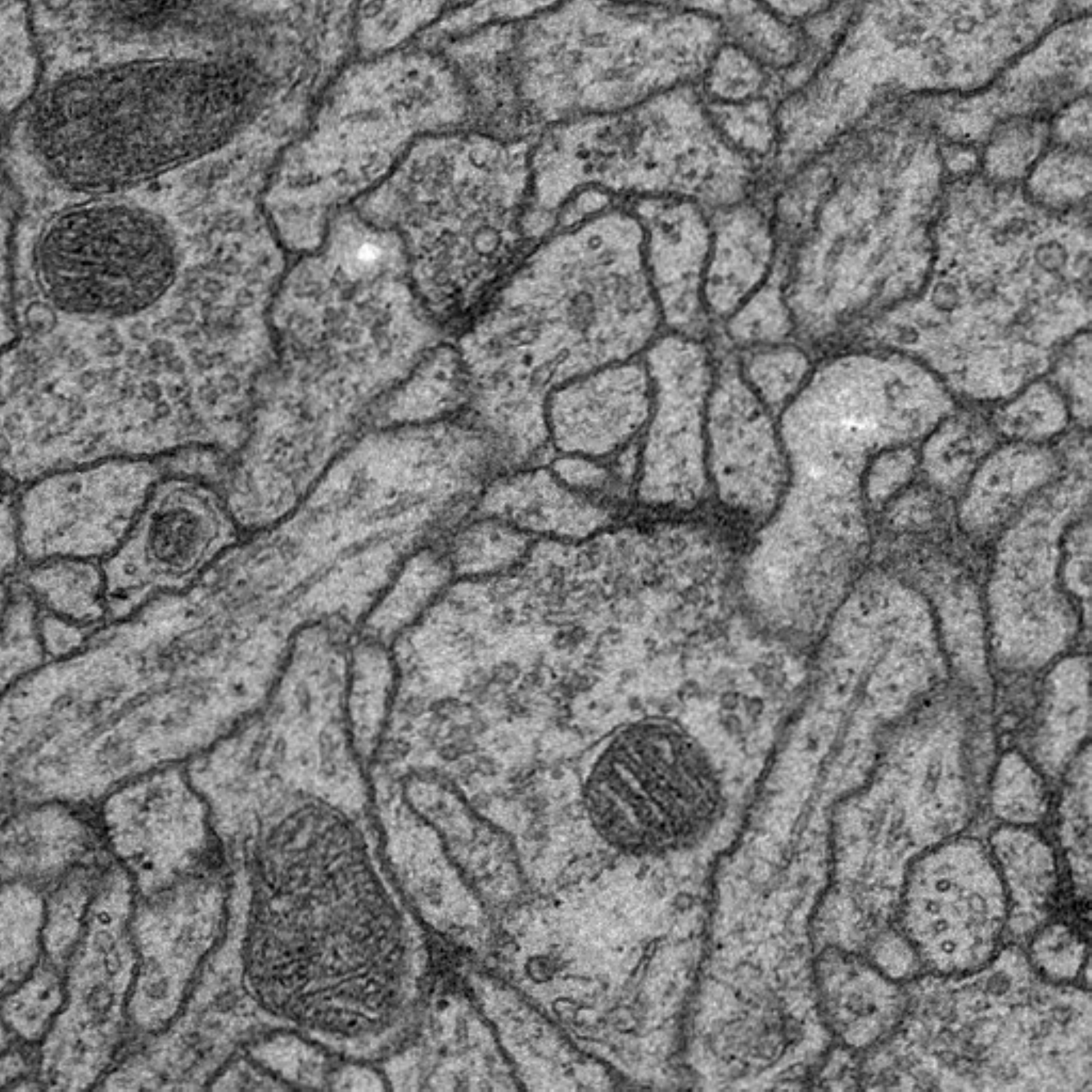}
                \caption{Input}
                \label{fig:T1}
        \end{subfigure}
        ~ %add desired spacing between images, e. g. ~, \quad, \qquad, \hfill etc.
        \begin{subfigure}[b]{0.3\textwidth}
                \includegraphics[width=\textwidth]{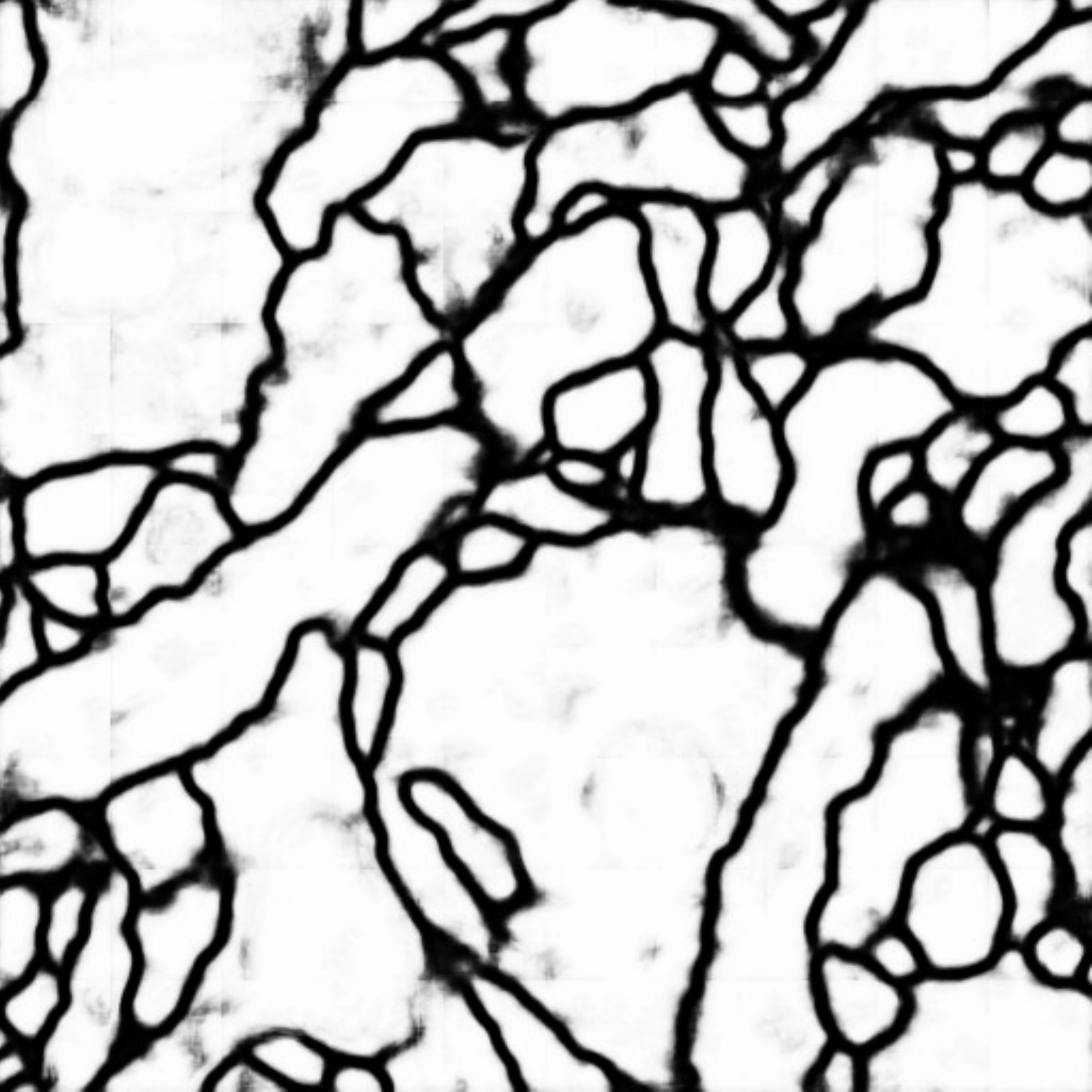}
                \caption{\abr{}}
                \label{fig:IR}
        \end{subfigure}
         \caption{Segmentation results on EM dataset (slice 26)}\label{fig:result-em}
\end{figure}

Membrane segmentation is evaluated through an online system provided by the ISBI 2012 organizers. 
Comparisons to other methods are reported in Table~\ref{tab:result-em}. 
The teams IDSIA and DIVE provide membrane probability maps for each pixel, like our method. 
These maps are adapted by the post-processing technique of the teams SCI~\cite{Liu14},
which directly optimizes the rand error (DIVE-SCI (top-1) and IDSIA-SCI (top-2)); 
this is most important in this particular segmentation task.
Without post-processing, \abr{} networks outperform other methods in rand error, 
and are competitive in wrapping and pixel errors. 
Of course, performance could be further improved by applying post-processing techniques.
Figure~\ref{fig:result-em} shows an example segmentation result.

\subsection{MR Brain Segmentation}
%dataset
%The dataset was provided by the ISBI 2015 workshop on Neonatal and Adult MR Brain Image Segmentation \cite{MICCAI15}.
%three error metrics
\paragraph{Evaluation metrics} The results are compared based on the following three measures:
\begin{itemize}
\item The DICE overlap (DC)~\cite{Dice1945}: spatial overlap between the segmented volume and ground truth
\item The modified Hausdorff distance (MD)~\cite{Huttenlocher1993}: 95th-percentile Hausdorff distance between the segmented volume and ground truth
\item The absolute volume difference (AVD)~\cite{Babalola2009}: the absolute difference between segmented and ground truth volume, normalized over the whole volume.
%juergen don't understand: the ratio of a difference???
\end{itemize}
%results
\paragraph{Results} 
\begin{table}[t]
\centering
\caption{The performance comparison on MR brain images.}
\setlength{\tabcolsep}{2.5pt}
\begin{tabular}{l|ccc|ccc|ccc|c}
\toprule
Structure    	& \multicolumn{3}{c|}{GM}  				& \multicolumn{3}{c|}{WM}  	& \multicolumn{3}{c|}{CFS}  &  \\
%\hline \\ [-1.7ex]
\cmidrule(r){1-10}
\multirow{2}{*}{Metric}   & DC & MD & AVD					& DC & MD & AVD 				& DC & MD & AVD & Rank \\
						 & (\%) & (mm) & (\%)			& (\%) & (mm) & (\%) 		& (\%) & (mm) & (\%) & \\

%    						 & ($10^{-1}$) & Warping Err. & Pixel Err.\\
%\hline \hline \\ [-1.7ex]
\midrule
BIGR2			& 84.65  		& 1.88 			& 6.14   		& 88.42			& 2.36			& \textbf{6.02} 	& 78.31	& 3.19	& 22.8 & 6\\
KSOM GHMF		& 84.12   		& 1.92  			& \textbf{5.44}  & 87.96			& 2.49			& 6.59 			& 82.10	& 2.71	& 12.8 & 5\\
MNAB2			& 84.50   		& 1.69  			& 7.10   		& 88.04			& 2.12			& 7.73 			& 82.30	& 2.27	& 8.73 & 4\\
ISI-Neonatology	& \textbf{85.77} & \textbf{1.62}	& 6.62   		& 88.66			& \textbf{2.06}	& 6.96 			& 81.08	& 2.66	& 9.77 & 3\\
UNC-IDEA			& 84.36    		& \textbf{1.62}  & 7.04   		& \textbf{88.69}	& \textbf{2.06}	& 6.46 			& 82.81	& 2.35	& 10.5 & 2\\
%\specialrule{.1em}{.05em}{.05em}
\midrule
\textbf{\abr{}}	& 84.82    		& 1.69 			& 6.77   		& 88.33			& 2.07	& 7.05 	& \textbf{83.72}	& \textbf{2.14}	& \textbf{7.10} & \textbf{1}\\
\bottomrule
\end{tabular}
\label{tab:result-mr}
\end{table}

MR brain image segmentation results are evaluated by the ISBI NEATBrain15 organizers~\cite{MICCAI15} who provided the extensive comparison to other approaches on \url{http://mrbrains13.isi.uu.nl/results.php}.
Table~\ref{tab:result-mr} compares our results to those of the top five teams. 
The organizers compute nine measures in total and rank all teams for each of them separately. 
These ranks are then summed per team, determining the final ranking (ties are broken using the standard deviation). 
\abr{} leads the final ranking with a new state-of-the-art result and outperforms other methods for CFS in all metrics.

We also tried regularization through dropout~\cite{srivastava14}. 
Following earlier work~\cite{dropout14}, the dropout operator is applied 
only to non-recurrent connections (50\% dropout on fully connected layers and/or 20\% on input layer).
However, this did not improve performance, but made the system slower.

\section{Conclusion}

Since 2011, GPU-trained max-pooling CNNs 
have dominated classification contests~\cite{ciresan:2011ijcnn,Krizhevsky:2012,zeiler2013}
and segmentation contests~\citep{ciresan2012nips}.
MD-LSTM, however, may pose a serious challenge to such CNNs,
at least for segmentation tasks. Unlike CNNs, MD-LSTM has 
an elegant recursive way of taking each pixel's entire spatio-temporal context into account,
in both images and videos. Previous MD-LSTM implementations, however, could not exploit
the parallelism of modern GPU hardware. This has changed through our work presented here.
Although our novel highly parallel \abr{} has already achieved state-of-the-art segmentation
results in challenging benchmarks, we feel we have only scratched the surface of what 
will become possible with such \abr{} and other MD-RNNs. 

\section{Acknowledgements}
We would like to thank Klaus Greff and Alessandro Giusti for their valuable discussions, and Jan Koutnik and Dan Ciresan for their useful feedback.
We also thank the ISBI NEATBrain15 organizers~\cite{MICCAI15} and the ISBI 2012 organisers, in particular Adri\"{e}nne Mendrik and Ignacio Arganda-Carreras.
Lastly we thank NVIDIA for generously providing us with hardware to perform our research.
This research was funded by the NASCENCE EU project (EU/FP7-ICT-317662).

\setlength\bibitemsep{0pt}
\printbibliography

\end{document}